# Discoverability in Satellite Imagery: A Good Sentence is Worth a Thousand Pictures

David Noever, Wes Regian, Matt Ciolino, Josh Kalin, Dom Hambrick, Kaye Blankenship*
PeopleTec, Inc. and US Army Space Missile Defense Command (SMDC)*, Huntsville, AL, USA
Corresponding author: david.noever@peopletec.com

## Abstract

Small satellite constellations provide daily global coverage of the earth's land mass, but image enrichment relies on automating key tasks like change detection or feature searches. For example, to extract text annotations from raw pixels requires two dependent machine learning models, one to analyze the overhead image and the other to generate a descriptive caption. We evaluate seven models on the previously largest benchmark for satellite image captions. We extend the labeled image samples five-fold, then augment, correct and prune the vocabulary to approach a rough min-max (minimum word, maximum description). This outcome compares favorably to previous work with large pre-trained image models but offers a hundred-fold reduction in model size without sacrificing overall accuracy (when measured with log entropy loss). These smaller models provide new deployment opportunities, particularly when pushed to edge processors, on-board satellites, or distributed ground stations. To quantify a caption's descriptiveness, we introduce a novel multi-class confusion or error matrix to score both human-labeled test data and never-labeled images that include bounding box detection but lack full sentence captions. This work suggests future captioning strategies, particularly ones that can enrich the class coverage beyond land use applications and that lessen color-centered and adjacency adjectives ("green", "near", "between", etc.). Many modern language transformers present novel and exploitable models with world knowledge gleaned from training from their vast online corpus. One interesting, but easy example might learn the word association between wind and waves, thus enriching a beach scene with more than just color descriptions that otherwise might be accessed from raw pixels without text annotation.

## 1. Introduction

Annotating images with short text descriptions or captions provides one method to create and automate tagging for large image repositories [Andres, et al., 2012]. Using keywords, subsequent text searches then can find and group similar images, thus rendering entire image collections as both indexable and discoverable [Blanchart, et al., 2010]. Similarly, with an effective caption generator specialized for satellite imagery, one can catalogue and inventory large areas of the planet and assist land use planners or first responders during a disaster recovery effort [Bratasanu, et al., 2010; Kyzirakos, et al., 2014]. To streamline this task, one needs to build a deep image search engine [Mao, et al., 2018], one which characterizes pixel features in words, extracts metadata from potentially millions of overhead images, and returns queries either for nearest matches in keywords or similar imagery.

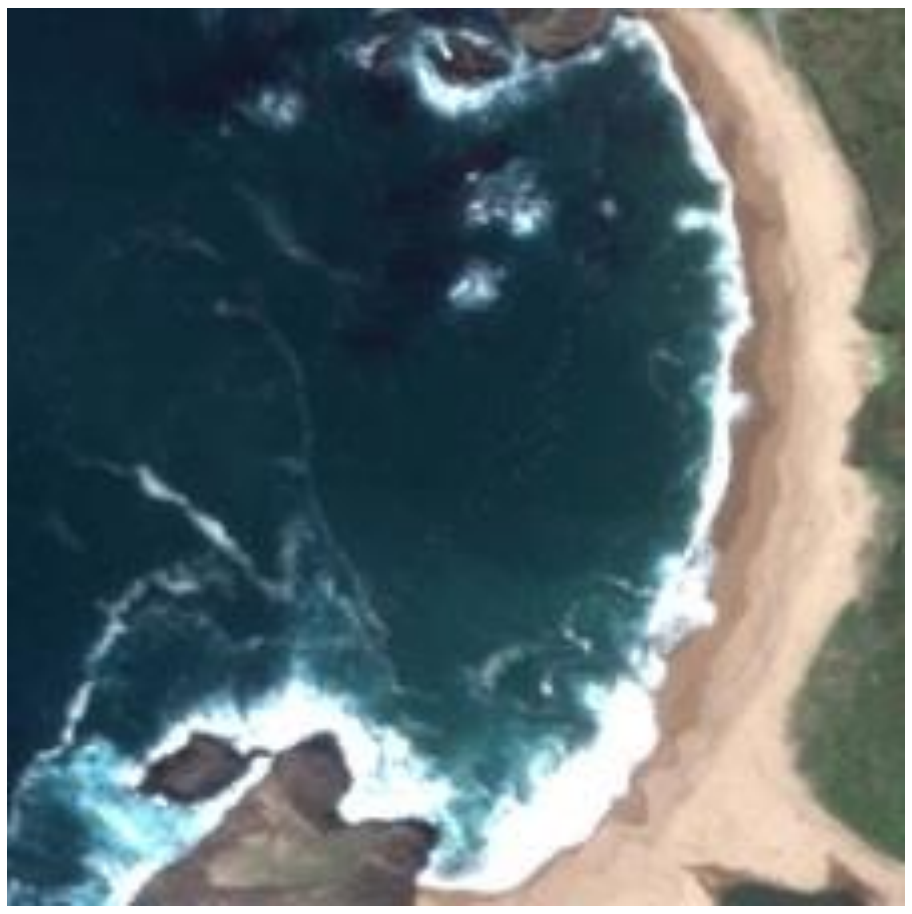

*Figure 1. Example Overhead Imagery to Caption. Human annotators label this training image as: "Yellow ribbon beach is between green trees and dark green ocean with white waves."*

*1.1 Broad Motivation.* To render the first functional map of the world, many researchers have advocated for a combination of automated labels with overhead imagery [Christie, et al., 2018; Demir, et al., 2018]. This global inventory would enable complex change detection not previously available. Furthermore, in cases when the annotation might occur on-board satellites, the automated captioning might guide future download choices and minimize data transfer of useless images. Opportunities for obtaining higher priority images would increase with pre-filtering of images degraded by poor camera optics, cloud cover, night, or broad open ocean [Yao, et al. 2016]. In two steps, we approach this satellite captioning problem for overhead images as firstly, recognizing all the objects in a complex satellite image, then secondly, describing all the discovered objects and their mutual relationships in a short sentence or annotation [Tanti, et al., 2018]. The final text serves as a captioned description that subsequently makes the image discoverable, shareable and searchable. For example, one might annotate an image like Figure 1 by combining scene recognition ("yellow beach") with all other objects relative to its adjoining scene parts ("next to a green forest"). This caption contrasts markedly with a generic model not specialized for overhead imagery, such as the Microsoft CaptionBot, which labels the same image as "I can't really describe the picture, but I do see black, looking, white" [Microsoft, 2019].

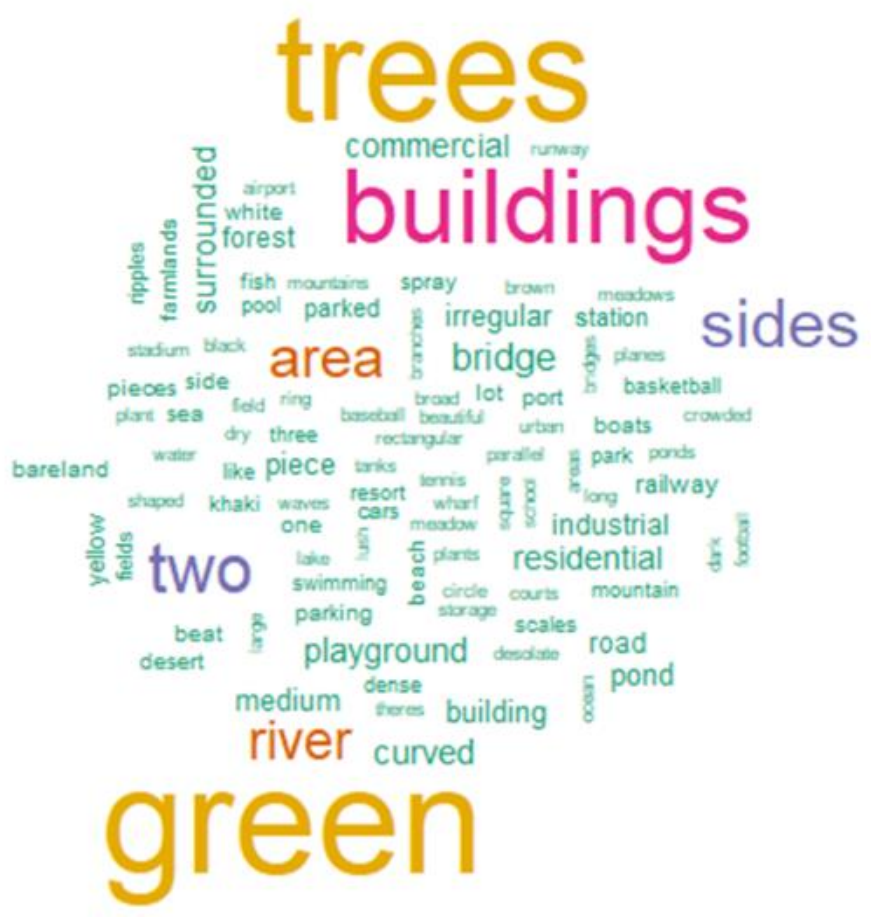

*Figure 2. Word cloud for frequently used annotation terms in RSCID dataset. The notable overabundance of colors, trees and buildings from human annotators indicate some repetitive labeling, particularly in residential land types.*

*1.2 Anticipated Outcomes.* For the present work, we apply multiple deep learning detection models to satellite images. We train these models using the technique of transfer learning, which leverages pre-training of feature extractors on much larger datasets, then extends the final image classification layer to previously undefined classes. Knowing the objects in the image, we generate captions using a recurrent neural network with long short-term memory [LSTM, see Gers, et al. 2000; Brownlee, 2017]. The overall method associates words and annotates complete caption sentences to those recognized objects [Luo, et al., 2011; Luo, et al., 2013]. Compared to previous work, our approach attempts to minimize the image model's overall size, thus making on-board satellite processing potentially viable, while also correcting and expanding the vocabulary traditionally included for large caption training steps. This work also explores the benchmark text vocabularies to include spelling corrections, synonyms and alternative sentence structures. Two unexpected outcomes of this re-examination of the initial captioning vocabulary follow from a built-in annotation bias, or a dominant sensitivity to color descriptions (which offer little new information over the raw red-green-blue pixels), and its inability to capture world knowledge that a human expert might offer such as describing the physical relationship between ocean white-caps and wind in an image. We explore the implications of these two outcomes more in depth in the Discussion section.

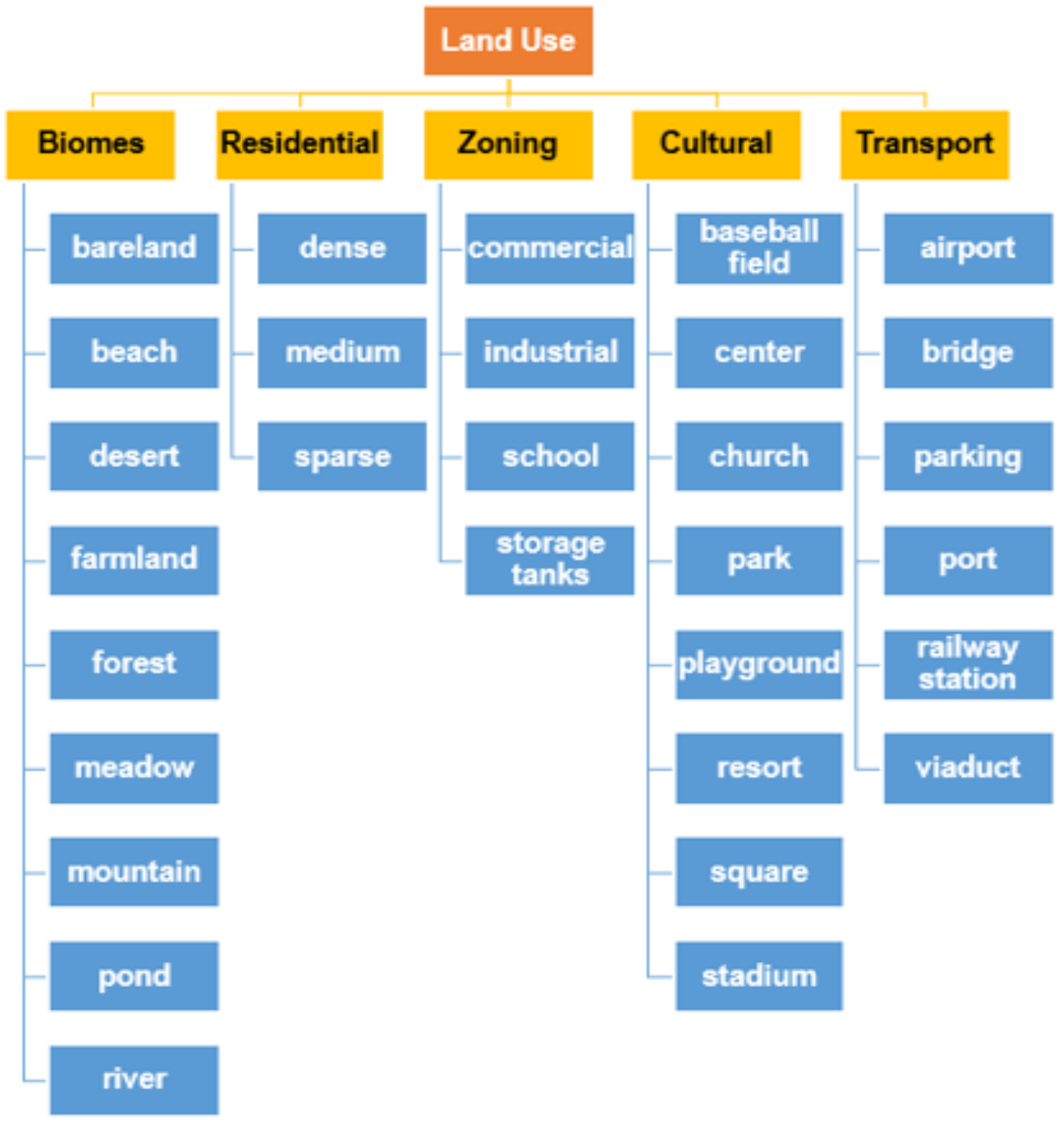

*Figure 3. Categories of characteristic satellite scenes highlighting land use in RSICD data, including biomes, residential, zoning, cultural and transport classes.*

*1.3. Previous Contributions.* Previous work [Cheng, et. al, 2017] has reviewed the challenges in classifying and captioning overhead imagery, noting particularly the lack of scene diversity (e.g. number of image classes). Various captioning benchmark efforts have appeared, some of which have offered a specialized earth-observation retrieval system based on the content of satellite imagery [Lu, et al. 2017] or have assembled a combined image and text dataset called the Remote Sensing Image Captioning Dataset (RSICD). To balance and diversify the objects recognized and described in previous captioning examples, the RSICD dataset reduces the relative number of included residential scenes. The authors suggest this residential imbalance biases previous captioning tasks, such as UCM [Yang, Newsam, 2010] and Sydney [Zhang, et al., 2019] datasets. While diversifying the available image classes (e.g. Figure 3), the RSICD authors follow a similar captioning format by providing five different sentences describing each image. They note that the caption diversity depends on two key factors, both how the five sentences differ from each other for a given image (textual depth, as shown in Figure 2) and how the different images differ in their respective captioning choices (image breadth, as shown in Figure 3). Both diversity types (the intra- and inter-image changes) may shape the trained algorithm's ability to generalize to new test images [Dumitru, et al., 2014; Espinoza-Molina, et al., 2013].

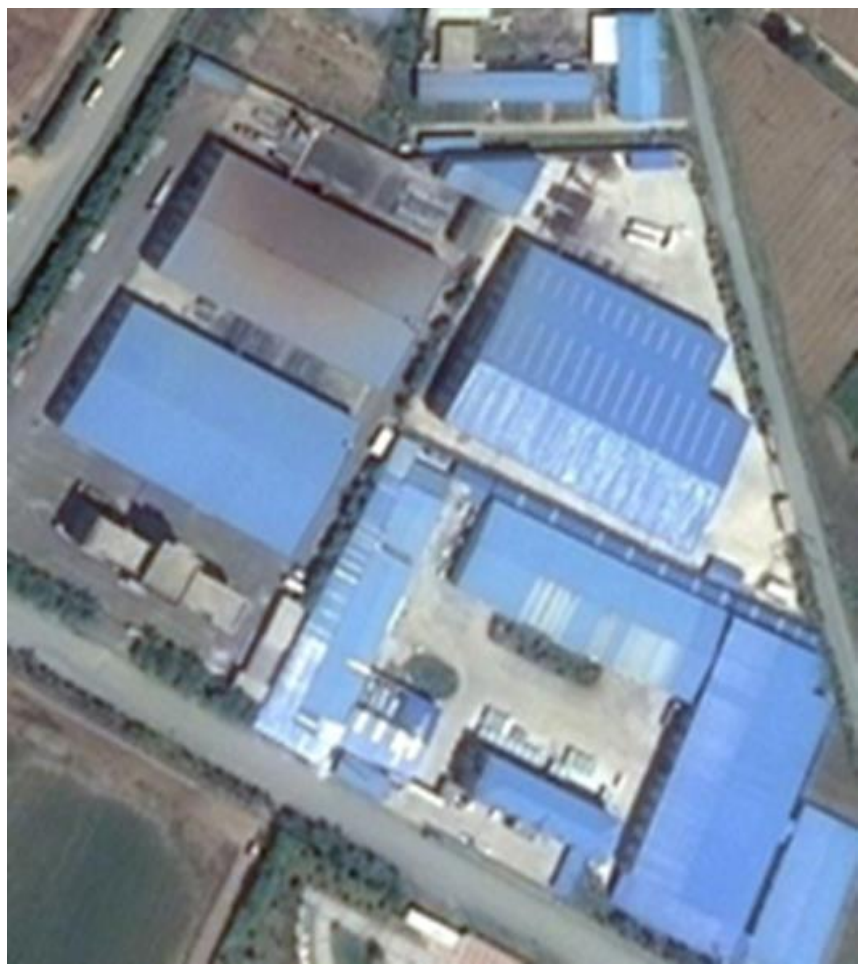

*Figure 4 Comparison captions from Microsoft's generic caption bot and the present specialized satellite version. The generic bot describes the image as, "I think it's a closeup of a building," where the satellite training set specializes to: "many buildings and some green trees are in an industrial area."*

*1.4 Original Contributions.* For the present work, we extend the RSCID dataset in three key ways. First. we supplement satellite image data [e.g. SIRI-WHU, Zhao, et. Al, 2015 and xView, Lam, et al, 2018]. Secondly, we deploy alternative pre-trained photo models, then modify the underlying captioning vocabulary. The latter attempts to cleanse and augment the overall vocabulary. Finally, we investigate how search discoverability might enable analysts to query large image repositories using both similar image and keyword matches [Shi, et al., 2017]. These results seek to discover if implementing such automated methods for assigning complex image metadata *in situ* might assist future satellite image analysts.

## 2. Methods

The research plan includes comparing many different image models to identify the best one for satellite images, then contrasting strategies for captioning them, some of which correct previous vocabulary shortcomings and others of which expand the baseline diversity of annotations. The initial RSICD dataset [Lu, et al., 2017] consists of 10,921 satellite images broadly grouped into 30 characteristic scenes. RSICD particularly highlights land uses such as residential, urban or agricultural classes. To understand the image clusters, we group the scenes into five major categories as shown in Figure 3, including land biomes, residential density, zoning, cultural and transport-related classes. It does not sub-class some satellite cases typically included elsewhere like roads, construction, and other change detection scenes used for damage assessment from space.

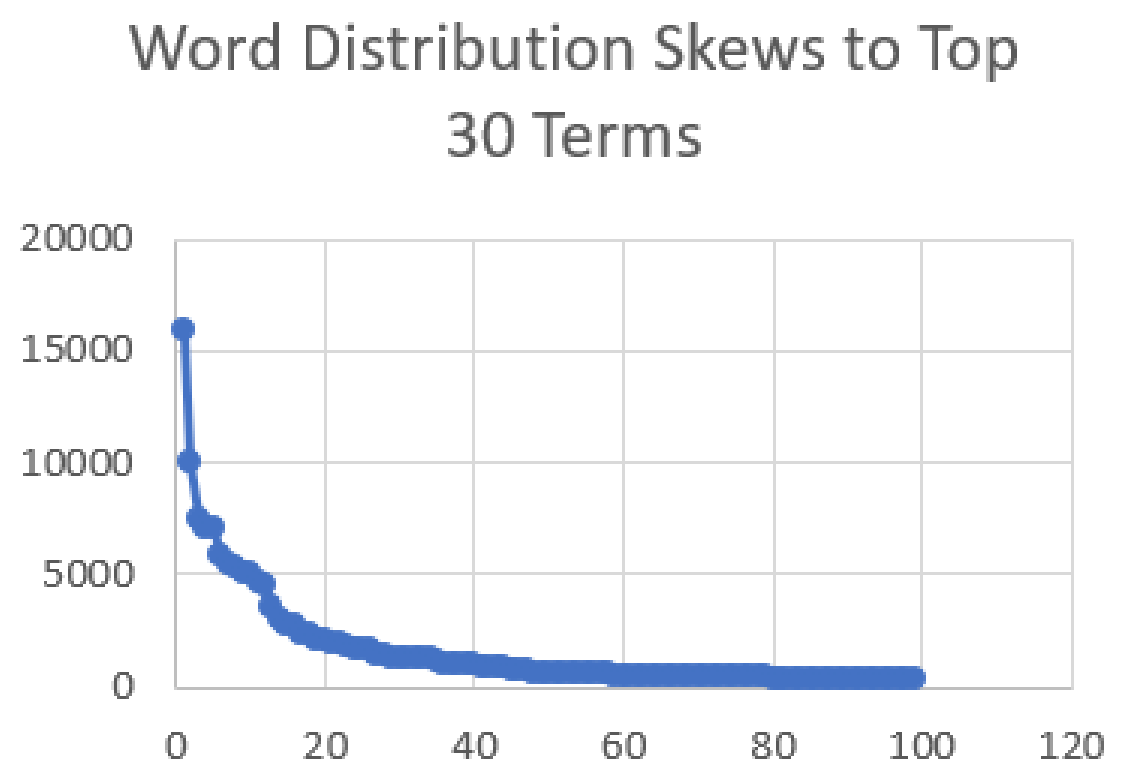

*Figure 4. Word frequency distribution in RSICD skews to the top 30 terms and color-related adjacency annotations such as "green trees next to a blue roof".*

*2.1. Dataset Image Descriptions.* Each satellite image is a small overhead chip, 224 x 224 pixels, collected at varying resolutions and sourced from overhead mapping services such as Google or Baidu. The dataset does not try to specify either the resolution in ground sample distance (GSD), or lineage by camera or satellite services. Each image carries five human-annotated captions, a high proportion (60+%) which include duplicate annotations. Thus, in total, the text portion of RSICD includes more than 50,000 sentences, 239,765 words, and the book equivalent of a 2100-page manuscript describing the 30 scene classes shown in Figure 3.

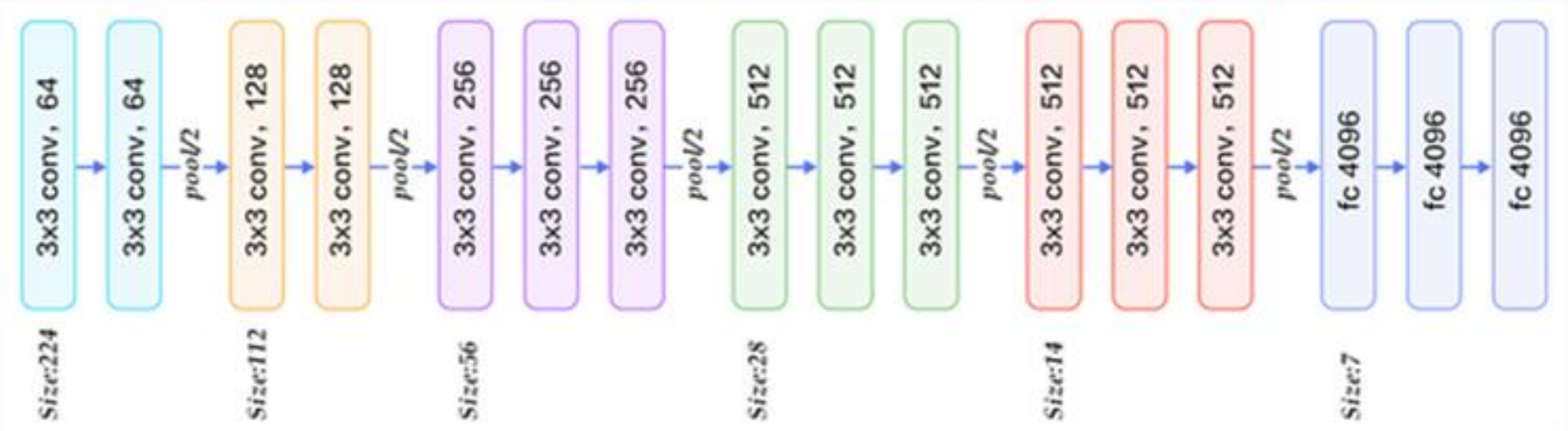

*Figure 5. VGG-16 Layered Convolutional Neural Network Architecture.*

*2.2. Dataset Text Descriptions.* A key aspect of RSICD's descriptive richness derives from its baseline size with 2643 unique words. However, while this text labeling represents a massive machine learning enterprise, it nevertheless can be both pruned and augmented to address some of its challenges. For example, while having human annotators produce 50,000 descriptions when viewing satellite imagery represents an undertaking, around sixty percent of the descriptions identically duplicate the others in the same series of five for each image (18,185 unique sentences, or 36,413 duplicates). It's unclear if this general copying strategy helps the ultimate task of captioning previously unseen test images, or whether the authors sought to over-sample and balance the training text data. Furthermore, we identified a substantial amount of mis-spellings (14%), or broken syntax, such as "c shape" or "t road" treated as two distinct words, when in context, annotators might have better described "C-shaped" as a single token.

| Models and Cross Entropy Loss | Training | Validation |
|---|---|---|
| **VGG19** | 1.815 | 1.891 |
| **VGG16** | 1.806 | 1.882 |
| **NASNetMobile** | 1.571 | 1.891 |
| **VGG19 + Corrected Caption** | 2.314 | 2.559 |
| **VGG19 + Synonym** | 3.004 | 2.896 |
| **NASNetMobile + Caption** | 2.268 | 2.508 |
| **NASNetMobile + Synonym** | 2.949 | 2.86 |

**Table 1. Model entropy loss for different vocabularies**

*2.3. Text Limitations.* As illustrated in Figure 4, the caption vocabulary follows a typical fat-tail word distribution that rapidly decays beyond the most frequent terms. In other contexts, this word commonality might make up a stop-word list. For example, half the total vocabulary (123,153 instances) include just the top thirty frequently used terms. In addition to the lack of descriptive diversity at the top rank of the distribution, there is a converse problem of many rare words at the long-end; for instance, 42% of the annotators' vocabulary are only used once (1409 of 3321 unique terms). At face value these initial limitations can be corrected by either cleansing with spell checking, pruning rare one-time uses with synonyms, or augmenting the overall vocabulary size with synonyms to reduce duplicates.

| Model | BLEU-1 | BLEU-2 | BLEU-3 | BLEU-4 |
|---|---|---|---|---|
| **VGG19** | 0.651538 | 0.480542 | 0.411383 | 0.304904 |
| **Lu, et al (Multimodal Best)** | **0.50037** | **0.3195** | **0.23193** | **0.17777** |
| **Lu, et al (Attention Best)** | **0.68968** | **0.54523** | **0.44701** | **0.3725** |
| **VGG16** | 0.659364 | 0.477588 | 0.405269 | 0.293555 |
| **NASNetMobile** | 0.650177 | 0.474976 | 0.405702 | 0.297666 |
| **VGG19 + Corrected Captions** | 0.541227 | 0.338149 | 0.268726 | 0.169482 |
| **VGG19 + Synonym** | 0.414955 | 0.254443 | 0.202806 | 0.119558 |
| **NASNetMobile + Caption** | 0.542573 | 0.338135 | 0.263581 | 0.162791 |
| **NASNetMobile + Synonym** | 0.51114 | 0.311813 | 0.246116 | 0.152852 |

**Table 2. BLEU Scores for different models and vocabularies**

*2.4. Text Augmentation Strategies.* We initiated all three strategies of textual augmentation [Wei, et al., 2019] and scored the results as both the models' ability to learn the new nomenclature correctly (entropy loss) and the captioned correspondence to a reference sample (Table 2, Bilingual Evaluation Understudy score, or BLEU-1 to BLEU-4). Higher BLEU scores mean a greater match between generated and reference captions, tending toward a limit of 1 [Papineni, et. Al, 2002]. Lower scores represent a higher divergence (akin to temperature in other language generation models). In a position-independent way, BLEU shows the number of (n-gram) matches between candidates and reference. The mean test BLEU for other benchmarks, like Flickr8k, are typically BLEU=0.37 (see Vinyals, et al. 2015). In summary, we initially divided 10,921 RSICD images (224x224) with 50,000 captions into a traditional split: 70% training, 15% development, 15% testing (N=1528 images). We score both the image and text models simultaneously, with images scored by entropy loss and the captioning quality assessed by BLEU (generated text compared to reference).

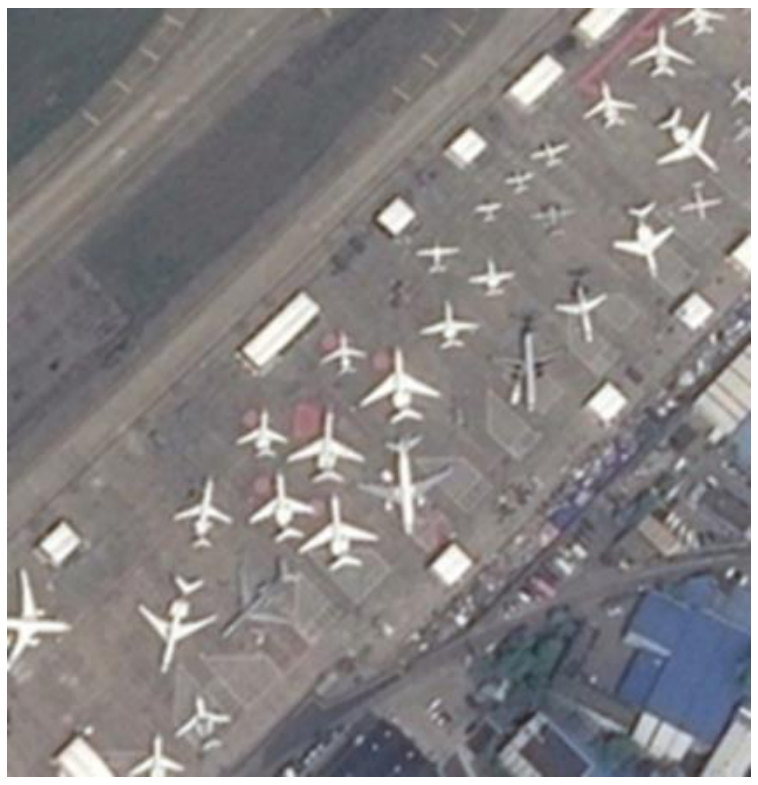

*Figure 6. Airport scene from RSCID, useful for comparing the different caption generation strategies,*

*2.5. Large and Small Image Models for Feature Extraction.* Our approach to the image model highlights mainly the size and architecture of each approach within traditional transfer learning, where the model is pre-trained except for fine-tuning on the final object classification layers. We initially applied VGG-16 and VGG-19 (Visual Geometry Group, Oxford), which stands out among other image classification networks because of its architectural simplicity [Simonyan, et al., 2014]. Only two building blocks are needed for VGG models, a 3x3 convolution and 2x2 pooling layer, throughout the entire network. One drawback of these simplified block choices however highlights the large network sizes needed to get good classification results, with either 16 or 19 layers as shown in Figure 5.

To compare VGG to smaller models, we also applied the Neural Architecture Search Network (NASNetMobile), which ranks 1.2% better in top-1 accuracy than the best human-invented architectures, while also reducing computational demands by 28% from previous state-of-the-art models [Zoph, et al., 2018]. The interest in these small but accurate models stem from the needs of low-power, edge computing applications with on-board satellite processors.

| Model | Generated Image Caption for Example Image from Figure 6: Airport_245 |
|---|---|
| **VGG19** | many buildings are in two sides of river with bridge over it |
| **VGG16** | many buildings and some green trees are in two sides of railway station |
| **NASNetMobile** | **many planes are parked in an airport** |
| **VGG19 + Caption** | many green trees and some buildings are in two sides of railway station |
| **VGG19 + Synonym** | many green trees and several buildings are around an almost circle gray center building |
| **NASNetMobile + Caption** | **many planes are parked near terminal in an airport** |
| **NASNetMobile + Synonym** | **many planes are in an airport near several buildings and some green trees** |

**Table 3. Example captions for different models and vocabularies applied to examples**

3. **Results**

We examined seven different cases for caption generation based on varying the underlying feature extraction model and extending the training vocabulary, either using synonyms, corrected syntax or reduced word diversity. Table 1 summarizes the experimental design along with overall results for different models and vocabulary approaches.

*3.1 Promise of Smaller Image Models.* Without any alteration of the captioning vocabulary, the best satellite image model for classification in RSICD was NASNetMobile, which similarly was the smallest in model size. The best captioning outcome was the large VGG19 model, but not substantially greater than NASNetMobile. The success of this small image annotation model offers a potentially faster approach in resource-scarce environments typically expected for edge computers.

| | airport | beach | desert | forest | harbor | railway | river | stadium |
|---|---|---|---|---|---|---|---|---|
| airport | 32 | 0 | 0 | 0 | 1 | 2 | 0 | 1 |
| beach | 0 | 372 | 13 | 0 | 0 | 0 | 9 | 0 |
| desert | 0 | 5 | 372 | 0 | 0 | 0 | 1 | 0 |
| forest | 0 | 0 | 22 | 446 | 0 | 0 | 3 | 0 |
| port | 17 | 14 | 0 | 0 | 364 | 1 | 0 | 1 |
| railway | 150 | 2 | 0 | 0 | 6 | 276 | 16 | 6 |
| river | 84 | 83 | 29 | 24 | 2 | 99 | 451 | 18 |
| stadium | 0 | 0 | 0 | 0 | 1 | 0 | 0 | 33 |
| Attributes | | | | | | | | |
| trees | 491 | 274 | 84 | 658 | 242 | 520 | 543 | 668 |
| white | 0 | 337 | 13 | 0 | 0 | 0 | 5 | 1 |
| yellow | 0 | 71 | 435 | 7 | 0 | 0 | 4 | 0 |

*Figure 7. Caption generator confusion matrix for NWPU test dataset with known scenes correctly labeled by caption mentions. The green diagonal shows correctly generated captions compared to the known NWPU scene labels. The lower table shows the descriptive attributes surrounding the overall scene annotations, such as white beach, yellow desert and trees in the forest.*

*3.2 Comparison to Previous Results.* To compare the quality of generated and reference captions, Table 2 shows BLEU scores for each captioning and model tested. Compared to Lu, et al. (2017), the BLEU scores shown in Table 2 exceed their best sequences for multimodal methods on RSICD. Harvesting just the best performers from their RNN and LSTM series (VLAD), we can compare the effects of variations in both the image and text models. The present version of transfer learning from the pretrained but large image model (VGG 19) and LSTM/RNN for its associated text closely correspond to Lu, et al. (2017). The much reduced NASNetMobile model similarly ranks highly compared to the best previous models with attention but at 100-fold smaller model size both in RAM and on disk.

| | airplane | beach | bridge | forest | parkinglot | railway |
|---|---|---|---|---|---|---|
| airport | 476 | 0 | 41 | 0 | 1 | 34 |
| beach | 0 | 445 | 4 | 0 | 0 | 5 |
| bridge | 48 | 2 | 1114 | 7 | 0 | 194 |
| forest | 0 | 2 | 311 | 683 | 0 | 0 |
| parking | 128 | 0 | 47 | 0 | 658 | 11 |
| railway | 4 | 0 | 141 | 0 | 0 | 124 |
| Attributes | | | | | | |
| river | 68 | 12 | 1517 | 11 | 0 | 236 |
| road | 59 | 2 | 446 | 32 | 38 | 63 |
| sea | 0 | 691 | 16 | 0 | 0 | 9 |
| trees | 154 | 32 | 2665 | 743 | 111 | 406 |
| waves | 0 | 87 | 0 | 0 | 0 | 0 |
| white | 0 | 359 | 4 | 0 | 0 | 4 |
| yellow | 0 | 176 | 51 | 0 | 0 | 4 |

*Figure 8. Caption generator confusion matrix for xView test dataset with known scenes correctly labeled by caption mentions. The lower table shows the descriptive attributes surrounding the overall scene annotations, such as white beach, yellow desert and rivers associated with bridges.*

*3.3 Comparative Example Between Model Cases.* To illustrate the alternate captions that each model and vocabulary can generate, we excerpt the description to a common airport scene (image Airport_245 from RSCID). Table 3 shows that in this case, the VGG image models incorrectly identify the airport as a railway station or bridge, while the NASNetMobile gets increasingly detailed and nuanced descriptive powers around the correct airport scene. This example highlights that improved syntax, whether corrected spelling or synonym diversity, does provide a richer descriptive vocabulary once the image model captures the key overhead objects of interest and orients them relative to each other. For instance, the baseline classification for many planes transforms after text augmentation steps to include other key potential landmarks, such as the terminal building, then finally identifying several buildings and green trees outside of the central tarmac identification [Wang, et al., 2019].

*3.4 Challenge to Diversify the Vocabulary.* The effect of augmented and corrected captions for training shows up as lower BLEU scores and higher image entropy losses. As one might expect for deep nets with a large set of tunable parameters, reducing the vocabulary can improve the memorization of input captions and lead to an apparently more accurate caption compared to a reference description. With that caveat, it's still not clear that one wants to introduce mis-spellings and repetition as an intentional effect from the outset of building a training set. These results highlight the tradeoffs in developing the most expressive description set possible with the least amount of resources, either computationally in model size or syntactically in vocabulary diversity. One wants a rich but accurate initial vocabulary [Li, et al., 2007; Zhao, et al., 2015], particularly one that corrects the residential bias of previous image collections but also minimizes repetitive use of color, buildings and trees when describing future overhead imagery. One would anticipate that no amount of textual augmentation is likely to improve a mis-classified image just by offering more detailed and incorrect annotation.

*3.5 Opportunities to Diversify and Augment the Test Dataset.* The airport example in Figure 6 and Table 3 can be generalized to much larger, well-known satellite datasets with extensive labeling efforts to see if the generated captions from those images alone can match with the known objects. For this case, we catalogued eight similar scenes from RCSID classes (airport, beach, desert, forest, port, railway, river and stadium), then deployed that trained model on the NWPU (Figure 7) and xView (Figure 8) satellite datasets in those same groups. In this way, the matrix shows a good image model for the strong diagonal scene match and the improved vocabulary model for the lower (heatmap) table associating the subject with its accompanying adjectives. This bootstrap testing approach is widely employed in other machine learning contexts but seems not to have been applied previously to the satellite captioning problem. This application offers a way out of the labor-intensive task of manual annotators. Example annotations performed on the large xView repository, which were previously not captioned, are shown in Figure 9. One notable outcome for the analyst is now the ability to discover, search and share millions of these images either by keyword queries or caption and pixel similarities [Cordeiro, et al., 2010].

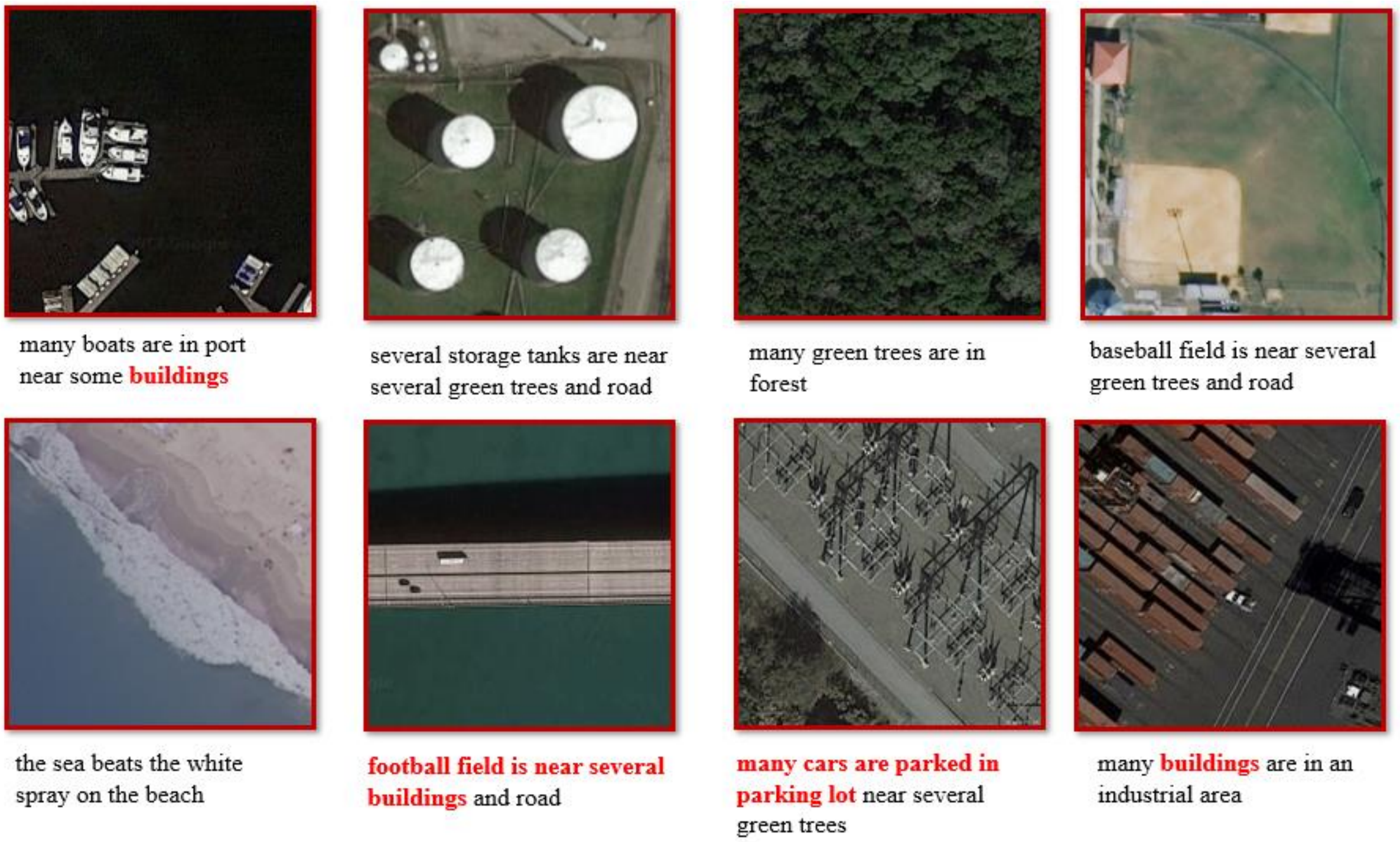

*Figure 9 Example captions trained with NASNetMobile on RSCID and tested on xView. The red annotations highlight errors in either the image model's ability to recognize the scene or objects outside of its class ontology such as transformer station or shipping.*

*3.6 Generalizing BLEU Scores with Novel Confusion Matrices.* It is worth noting that unlike the 10,000 images captioned by RCSID or 800 images captioned by NWPU (REmote Sensing Image Scene Classification (RESISC), created by Northwestern Polytechnical University (NWPU), [Cheng, et al., 2017]), the xView dataset alone collects bounding box detections on close to 1 million objects, many of which in image chips can be aggregated to form complex scenes like parking lots, airports and construction sites. As comparable ground truth, the "unknown" captions in xView cases can be confidently inferred because all the objects (yielding "nouns") are known along with their spatial relationships (yielding "adjectives" and "verbs"). Together this triple of subject-predicate-object [Yang, et al., 2011] constitutes a well-formed caption so that xView scales well to a large repository of discoverable and searchable images. An example query for xView, in this case, might simply ask, "Show me all the airports near a river bridge". After a monsoon or widespread flooding damage, that application might enable assessments of transport accessibility and availability by ground and air.

This approach generates a traditional multi-class confusion matrix, with the diagonal indicating the number of instances in which our trained caption generator can include the correct matching keywords for the actual scenes as labeled by either xView (Figure 8) or NWPU (Figure 7) object detectors. For instance, a true positive for a known xView airport scene should include a caption mentioning the term "airport", which then would be counted as correctly assigned or true positive in these captioning error matrices.

This method effectively isolates the overall subject-noun relationship one might hope for in a decent sentence generator, and largely evaluates the accuracy of the image model itself. In other words, at a minimum, a good beach scene description should mention the noun, beach. The more nuanced language of a human analyst might provide additional adjectival support, such as describing the beach as white or the desert as yellow. The lower heatmaps in both Figures 7-8 isolate the keywords found common in both the bounding box description and the generated text output from the image annotator. Note that the annotation has no prior knowledge of the existing xView object detections; each image is processed as a new test case then the output is compared to what the human labelers for xView also marked with boxes. For example, the lower half of Figure 8 shows that a beach scene from xView includes in both our annotations and in its own labels, the expected elements of "sea", "waves", "white" and "yellow". A reasonable caption thus could iterate variations on the short description: "A yellow beach next to the sea with waves".

## 4. Discussion

This research has tested seven new RNN-LSTM transfer-learning models for satellite image captioning. The models were trained initially on the RSCID dataset, then compared to both pruned and augmented vocabularies for better annotations. The image models included both large (VGG-16 and VGG-19) to compare with previous literature, then extends the work to smaller (NASNetMobile) models that can outperform in both accuracy (entropy loss), size and speed. The aim of reducing the image model size 100-fold is partly to explore captioning as an edge processing task on-board small satellites with limited processing power.

| Counts | RSCID | RSCID-Corrected | RSCID-Synonym |
|---|---|---|---|
| **Characters** | 3,045,229 | 1,083,812 | 4,813,070 |
| **Words** | 524,997 | 186,168 | 803,212 |
| **Unique Words** | 3,037 | 2,560 | 3,447 |
| **Complex Word %** | 7.70 | 8.07 | 10.39 |
| **Avg. Syllables / Word** | 1.42 | 1.40 | 1.48 |
| **Sentences** | 54,573 | 18,168 | 73,276 |
| **Avg. Words/ Sentence** | 9.62 | 10.25 | 10.96 |
| **Fog grade level** | 6.93 | 7.33 | 8.54 |
| **Flesch reading ease** | 77.32 | 77.71 | 70.82 |
| **Flesch-Kincaid level** | 4.86 | 4.97 | 6.10 |

**Table 4. Metrics for augmentation and pruning caption vocabularies**

*4.1 Effect of Vocabulary.* While BLEU scores go down with augmented vocabularies, this metric highlights comparisons to a reference sentence, which in our case, becomes less well-defined as the vocabulary gets larger and more diverse (see Table 4). The major effect of augmenting and pruning caption vocabulary is to raise both the word- and reading-complexity while either reducing or keeping relatively constant the total number of unique words. This tradeoff highlights the min-max strategy for creating the richest descriptions in the fewest words. A secondary effect is to raise (and lower) selectively the total caption count by 60 percent. By examining the example output qualitatively, the resulting large vocabulary appears to provide a more convincing caption, although all text generation models will fail in the absence of a good image model.

*4.2 Overall Summary* To make this point quantitatively, we increased the test images by more than a factor of five compared to the largest previous captioning task, then applied our trained annotator to generate descriptions and scored that output in multi-label confusion matrices. We believe this novel approach holds promise to leverage the large amount of existing satellite detection labels and locations. It also alleviates a bottleneck in the laborious task of human labelers required to generate the approximately 2100-pages of annotations published in RSCID. In summary, this work has introduced new image and text models, captioned a five-fold larger dataset for future work, and applied confusion matrices in a novel way to highlight caption success without needing a prior reference sentence for BLEU comparisons.

*4.3 Further Research Opportunities* When considering future approaches, current human annotators leave out much of their associated world knowledge. For instance, Lu, et al. (2017) offers annotation instructions to identify image parts in six words, without using compass directions (North), "There is…", or vague descriptions like tall, large, or many. However, even a young adult might note much more about the physical world, such that the beach scene in Figure 1 looks like "a windy day" or "Californian rocky beach". In either case, the better caption requires some insight into either physical association between waves and wind, or geo-location familiarity.

| Input | Back Translated English |
|---|---|
| Many trees behind a school bus | Many trees behind a school bus |
| Island next to crashing | Island with waves |
| Airport terminal with many planes | Airport terminal with many levels |
| many buildings are in two sides of river with bridge over it | Many buildings are on both sides of the river with a bridge over |
| many buildings and some green trees are in two sides of railway station | many buildings and green trees on both sides of the station |
| many planes are parked in an airport | parked many aircraft at the airport |
| many green trees and some buildings are in two sides of railway station | many green trees and buildings are on both sides of the station |
| many green trees and several buildings are around an almost circle gray center | lots of green trees and several buildings around a central building almost back to gray |
| many planes are parked near terminal in an airport | many aircraft are parked near an airport terminal |
| many planes are in an airport near several buildings and some green | many aircraft at an airport near several buildings and green trees |

*Figure 10. The back translation from English input through 3 intermediate translation APIs and then returned to English as automated augmentations*

Further promising work should augment captions by not only substituting synonyms or correcting phrasing as done here but also trying new methods such as passing each caption multiple times through language translators and back translation, which has proven to yield interesting results [Ma, 2019; Wei, et al. 2019]. For example, after passing a complex translation cycle (English-to-Spanish, Spanish-to-German, German-to-French, French-to-English), the initial caption, "Island next to crashing waves", simplifies and generalizes into "Island with waves". This strategy illustrated in Figure 10 provides automatic caption generalizations without changing their true labels.

Future work should also highlight alternate pre-trained image models (ResNet, U-Net, etc.) and text transformers (BERT and GPT variants) to improve the overall output [see Budzianowski & Vulić, 2019]. It is worth noting however that the model sizes may not be compatible with edge processors and will likely overfit and amplify many of the anomalies found here in existing annotation data, such as mis-spellings, incorrect tenses and repetitive labels.

**Acknowledgements.** The authors would like to thank the PeopleTec Technical Fellows program for encouragement and project assistance. This research benefited from support from U.S. Army Space and Missile Defense Command/Army Forces Strategic Command.